\documentclass{article}
\usepackage{float}

\usepackage{graphicx}
\usepackage{multirow}
\usepackage{longtable}
\usepackage{tabularx}
\usepackage{booktabs}
\usepackage{amsmath}
\usepackage{amssymb}
\usepackage{bm}
\usepackage[utf8]{inputenc}
\DeclareUnicodeCharacter{2212}{\textminus}
\usepackage{multirow}
\usepackage{array}
\usepackage{booktabs}

\usepackage[preprint]{neurips_2024}

\usepackage[utf8]{inputenc} 
\usepackage[T1]{fontenc}    
\usepackage{hyperref}       
\usepackage{url}            
\usepackage{booktabs}       
\usepackage{amsfonts}       
\usepackage{nicefrac}       
\usepackage{microtype}      
\usepackage{xcolor}         

\title{SegICL: A Multimodal In-context Learning Framework for Enhanced Segmentation in Medical Imaging}

\author{
    Lingdong~Shen$^{1,2\dag}$\quad
    Fangxin~Shang$^{3}$ \quad
    Xiaoshuang~Huang$^{3}$ \quad\\
    \textbf{Yehui~Yang}$^3$\thanks{Correponding author.}\quad
    \textbf{Haifeng~Huang}$^3$ \quad
   \textbf{ Shiming~Xiang}$^{1,2}$\footnotemark[1]\\[3pt]
    $^1$School of Artificial Intelligence, University of Chinese Academy of Sciences (UCAS) \\ 
    $^2$State Key Laboratory of Multimodal Artificial Intelligence Systems (MAIS), \\ 
    Institute of Automation, Chinese Academy of Sciences (CASIA) \\
    $^3$Healthcare Group, Baidu Inc.\thanks{Work performed during an internship at Baidu Inc.} \\[3pt]
    \small{\texttt{shenlingdong2022@ia.ac.cn}}\quad\\
    \small{\texttt{yangyehuisw@126.com}}\quad
    \small{\texttt{smxiang@nlpr.ia.ac.cn}}\quad
}

\begin{document}

\maketitle

\begin{abstract}
In the field of medical image segmentation, tackling Out-of-Distribution (OOD) segmentation tasks in a cost-effective manner remains a significant challenge.
Universal segmentation models is a solution, which aim to generalize across the diverse modality of medical images, yet their effectiveness often diminishes when applied to OOD data modalities and tasks, requiring intricate fine-tuning of model for optimal performance. 
Few-shot learning segmentation methods are typically designed for specific modalities of data and cannot be directly transferred for use with another modality.
Therefore, we introduce SegICL, a novel approach leveraging In-Context Learning (ICL) for image segmentation. 
Unlike existing methods, SegICL has the capability to employ text-guided segmentation and conduct in-context learning with a small set of image-mask pairs, eliminating the need for training the model from scratch or fine-tuning for OOD tasks (including OOD modality and dataset).
Extensive experimental demonstrates a positive correlation between the number of shots and segmentation performance on OOD tasks. The performance of segmentation when provided thre-shots is approximately 1.5 times better than the performance in a zero-shot setting.
This indicates that SegICL effectively address new segmentation tasks based on contextual information.
Additionally, SegICL also exhibits comparable performance to mainstream models on OOD and in-distribution tasks.
Our code will be released after paper review.

\end{abstract}

\section{Introduction}
\label{sec:intro}
Medical imaging segmentation is a crucial component of biomedical image analysis and has become a vital tool in medical diagnosis and health monitoring. 
It have evolved from initial CNN structures\cite{a42,zhou2018unet++,a3} to the vision transformer\cite{a4,a5,a6,a20}.
Preliminary research outcomes have also been achieved in areas such as efficient data utilization\cite{a7,a8} and the design of universal models\cite{a9,a10,a12}. 
With the advancements in this area, researchers seek a framework with strong generalization performance, low training and maintenance costs, and user-friendly.

In the field of medical imaging segmentation, enhancing model generalization performance on OOD tasks and reducing develop cost remains a challenge. Traditional methods such as Few-Shot Learning (FSL) aim to improve model generalization with a small number of shots \cite{a16,a17,a18}.
However, specific FSL models often excel only within specific modalities, and tend to underperform when applied to other types of medical imaging datasets. Semi-supervised learning approaches use data augmentation or pseudo-labeling techniques \cite{a8,a11,a13,a14,a19, a7,a15} to reduce the costs associated with data annotation. However, these methods usually require fine-tuning or retraining when facing to new tasks, which incurs additional costs and effort. ICL methods offers a promising alternative, as it can effectively handle Out-of-Distribution (OOD) tasks in a cost-efficient manner. It do not require retraining to adapt to new tasks, which significantly lowers maintenance costs. Despite their potential, existing ICL methods have primarily been explored in domains such as language tasks \cite{a63,a64,a65,a60,a61}, image generation \cite{a66,a67}, image segmentation and grounding \cite{a50,a70,a71}, multimodal understanding \cite{a72,a23,a74,a76,a62}, and embodied vision \cite{a75}. As far as we know, there have been limited attempts to apply ICL to fine-grained medical perception tasks like segmentation. From the perspective of user-friendly interaction, one notable method involves building a universal segmentation model based on the SAM\cite{a28}. This model uses geometric information such as points and boxes as prompts to guide segmentation. However, this approach can be less user-friendly if users lack the medical knowledge to accurately identify geometric locations of certain organs.

In summary, while FSL methods can somewhat alleviate OOD task issues, they struggle with cross-modal datasets. Semi-supervised methods help reduce development costs by minimizing data annotation requirements but still need fine-tuning or retraining for new tasks. Existing ICL methods show strong generalization and adaptability without retraining, yet they have not been adequately explored for fine-grained medical tasks.
\begin{figure}[H]
  \centering
  \includegraphics[width=1\textwidth]{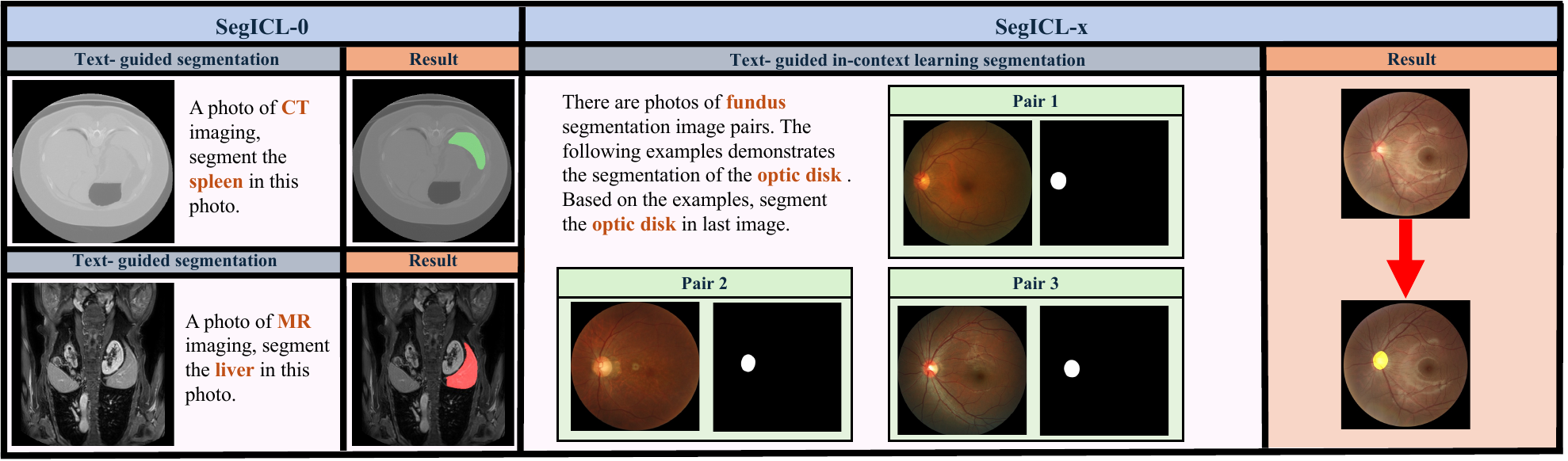}
  \caption{The example of inference pipeline of SegICL. The left part demonstrates SegICL can following  text instructions. The right part demonstrates SegICL can segment OOD data with a few image-mask pairs and text instructions.}
  \label{fig:1}
\end{figure}
To address these limitations, we propose SegICL, the first multimodal interactive in-context learning paradigm for medical segmentation, as shown in Fig.
\ref{fig:1}. SegICL addresses OOD tasks by performing ICL with few shots, making it suitable for segmenting various image modalities and avoiding the poor transfer performance of FSL. In terms of cost reduction and adaption of new tasks, SegICL adapts to new tasks with a training-free manner, thereby lowering maintenance and training costs associated with model fine-tuning or retraining. For user-friendliness, SegICL leverages text to guide the segmentation of target areas, which reduces the adaptation threshold for users.

Our contributions are as follows. Firstly, we propose a ICL paradigm for medical image segmentation, which can learn a in-context visual representation and follow multimodal instructions. Secondly, SegICL utilizes ICL to solve segmentation of OOD tasks, tackles region-of-interest segmentation through user-friendly instructions. Thirdly, extensive experiments have demonstrated that SegICL exhibits impressive performance across multiple medical image segmentation datasets, showcasing competitive capabilities comparable to other SOTA methods.

\section{Related work}
\label{sec:Related}
\subsection{Universal medical image Segmentation}
With the rise of universal vision models such as CLIP\cite{a29}, SAM\cite{a45} and SegGPT\cite{a50}, the field of biomedical image segmentation has also witnessed a surge in research on this topic. These studies can be classified into two types.
One method involves fine-tuning or replacing new modules based on existing general vision models (such as SAM)\cite{a46,a47,a48,a49,a10}. For example, SAMed\cite{a46} adopts a low-rank-based fine-tuning strategy (LoRA) on top of SAM, fine-tuning the SAM image encoder on labeled medical image segmentation datasets. Another example is Med-SA\cite{a47}, which differs from fine-tuning the SAM model by proposing the Medical SAM Adapter (Med-SA), incorporating domain-specific medical knowledge into the segmentation model. The other type involves training specialized universal medical segmentation models from scratch based on large-scale datasets. For instance, UniverSeg\cite{a9} utilizes a novel CrossBlock mechanism to generate accurate segmentation mask without requiring additional training, achieving generalization to new tasks.

Our approach differs from the methods mentioned above. The core of our proposed paradigm lies in a multimodal input-driven segmentation method that incorporates in-context learning based on textual instructions. None of the aforementioned methods can simultaneously achieve both text-guided segmentation and context-aware segmentation, which are the two core aspects of our approach.

\subsection{Learning from limited data samples}
Few-shot medical image segmentation\cite{a16,a17,a18,a22,a24,a38} is designed to efficiently address the scarcity of learning and generalizing from a limited number of shots. In the context of segmentation tasks, few-shot learning is introduced to segment unseen classes with the limited dataset. VQNet\cite{a14} introduces a VQ learning mechanism with Grid-Format VQ (GFVQ), Self-Organizing VQ (SOVQ), and Residual-Guided VQ (ROVQ) to medical MRI image segmentation. SE-Net\cite{a39} incorporate few-shot learning for segmenting abdominal organs in CT images. 

SegICL's motivation aligns with few-shot learning, aiming to learn OOD tasks or adapt to OOD datasets with minimal annotated data. However, these approaches exhibit poorer transfer performance, making them challenging to apply across modalities and tasks. 
Moreover, when confronted with OOD tasks from different modalities, they struggle to complete the segmentation accurately. In contrast, the proposed paradigm possesses the capability of multimodal ICL, enabling it to wrok without retraining when facing OOD tasks.

\section{Methodology}
\label{sec:Method}
\begin{figure}[H]
  \centering
  \includegraphics[width=0.98\textwidth]{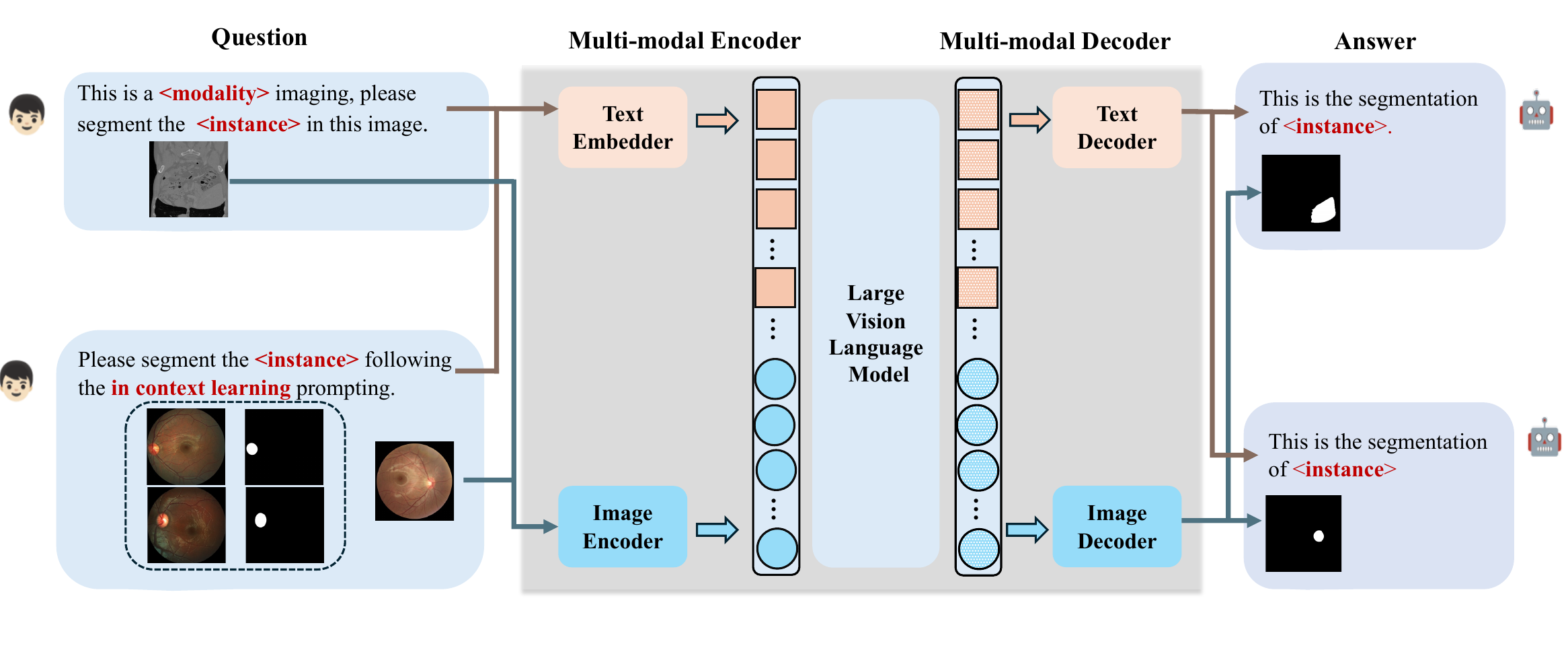}
  \caption{The overall structure of the SegICL paradigm.
  SegICL-0 represents zero-shot inference, while SegICL-x represents using x image-mask pairs as prompts for inference.}
  \label{fig:2}
\end{figure}
\subsection{In-Context Learning Paradigm of SegICL}
The learning paradigm of SegICL aims to utilize information from multimodal input, enabling to extract implicit data correlations in multimodal data. 
Additionally, it empowers the model with the ability to perform segmentation using text-guided methods.

\begin{equation}
  H = Enc(X_{img},X_{text})
  \label{eq:1}
\end{equation}
\begin{equation}
   S = Proj(H)
  \label{eq:2}
\end{equation}

\begin{figure}[H]
  \centering
  \includegraphics[height=7cm]{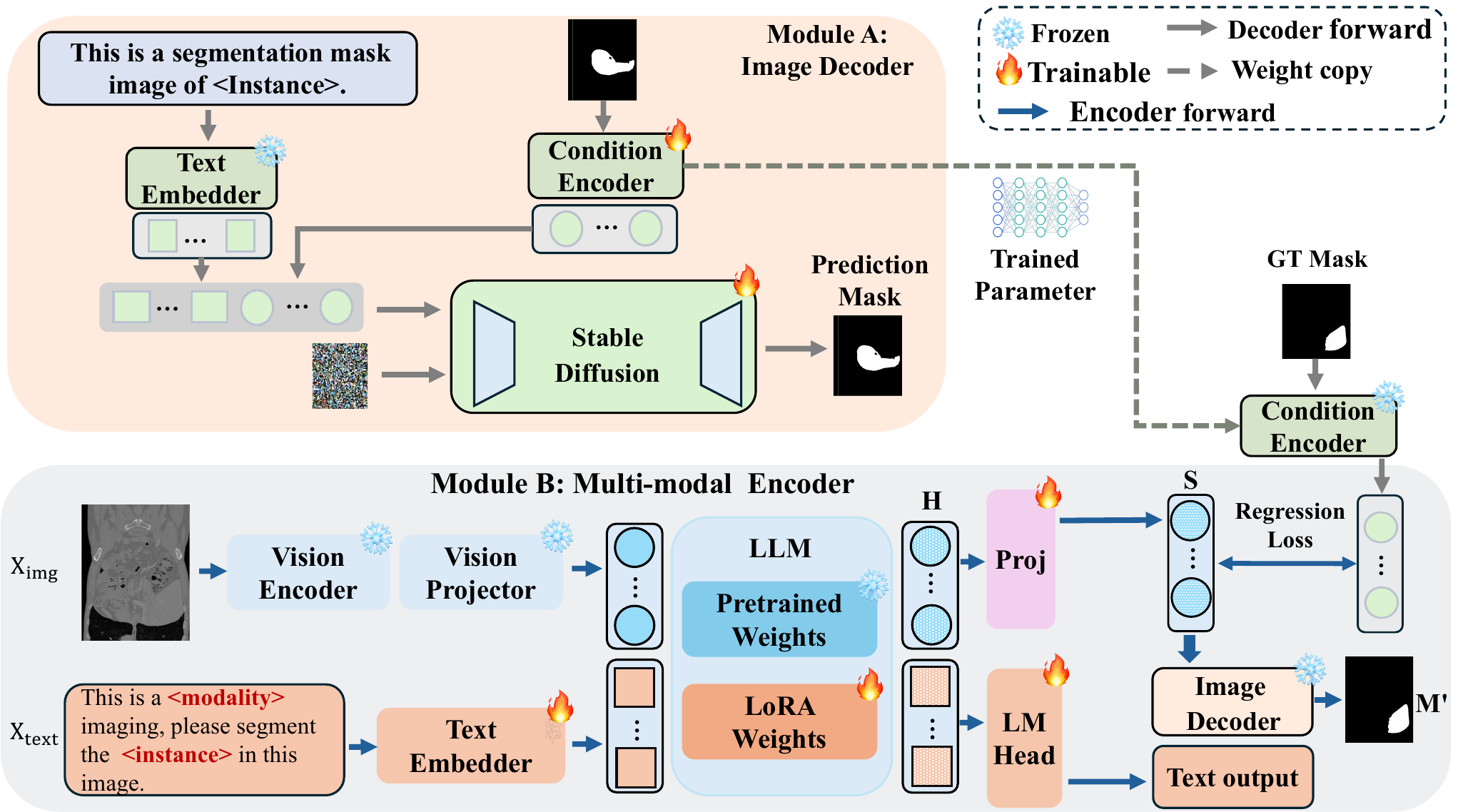}
  \caption{Training pipeline of SegICL. Module A is a image decoder, and Module B is a multimodal encoder. Interleaved multimodal text and image input undergoes encoding in Module B (get the hidden variable H ), and the projected data (get the result state S) is then passed to Module A for decoding, ultimately generating the corresponding mask.}
  \label{fig:3}
\end{figure}

The learning paradigm of SegICL (See Fig.\ref{fig:2}) can be represented as eq.\ref{eq:1}-eq.\ref{eq:3}: where $X_{img}$ and $X_{text}$ represents the input including images and texts respectively, $Enc$ is a large language model (LLM). After passing the multimodal input through $Enc$, it obtains a hidden variable ($H$). Then, using a projector to align the $H$ of the encoder with the feature space of the decoder, obtaining the encoded result state ($S$). Finally, the state is fed into the $Img_{Dec}$ for decoding, resulting in the final output mask. 
\begin{equation}
   M' = Img_{Dec}(S)
  \label{eq:3}
\end{equation}
The $Enc$ is a trainable function that accepts multimodal input and regresses to a hidden state vector. The $Img_{Dec}$ can be any image generation model, decoding the final prediction from above state. The $Enc$'s supervision signal comes from the regression loss between the encoded result state and the condition encoder (See Fig.\ref{fig:3}). The $Img_{Dec}$'s supervision signal comes from the regression loss between the generated image and the ground truth masks.
\subsection{Architecture}
Building upon the aforementioned learning paradigm, we propose the first universal text-guidence in-context learning medical segmentation framework SegICL. This framework (See Fig.\ref{fig:3}) use a LLM implementing the $Enc$ to model the input multimodal data. Additionally, it includes a diffusion model  ($Img_{Dec}$) to decode the state vector outputted by the encoder, yielding the final prediction. A MLP serves as the shared condition encoder to provide regression targets for the encoder model.

\begin{equation}
 M' =Img_{Dec}(Proj((Enc(X_{img},X_{text}))))
  \label{eq:4}
\end{equation}

The forward computation of the framework is represented by the equation above (see Fig.\ref{eq:4}). This framework initially takes arbitrary interleaved textual and visual data as input. After processing through a visual encoder and a text encoder, it generates interleaved tokens of text and image (See Fig.\ref{fig:3} module B). These tokens serve as input for the multimodal model, generating latent variables within a transformer structure rich in causal self-attention mechanisms. Since the encoder and decoder are decoupled, a projector is needed to map the model's feature space. After the tokens pass through the projector, a state vector is generated and made available for the decoder. The decoder model utilizes a diffusion model to ensure the quality of the generated images. When the input state vector is provided to the decoder model, it introduces random noise and uses the state vector as a condition to predict the output mask (See Fig.\ref{fig:3} module A). SegICL's $Img_{Dec}$ is trained using ground truth mask images. Its input includes random noise and a conditional vector (obtained by encoding mask images through the condition encoder). The output is a segmentation mask, with the goal of reconstructing the original image using the vector provided by the shared condition encoder.

\subsection{Multi-modal Encoder}
In contrast to existing multimodal large models\cite{a22,a23,a24,a25}, the multimodal encoder used in this paper aims to regress the supervision signals provided by the condition encoder. Therefore, a lightweight condition encoder is incorporated for supervised fine-tuning of the multimodal Encoder. 
\begin{equation}
\mathcal{L} = \frac{1}{n} \sum_{i=1}^{n} \left( \text{H}_i - \hat{\text{C}}_i \right)^2
\label{eq:6}
\end{equation}
The input $img \in \mathbb{R}^{H\times W \times 3}$ and $text \in \mathbb{R}^{\ Len \times 1}$  is processed by $Enc$ to obtain $M \in \mathbb{R}^{C_{hs} \times N_v}$, where H represents a multimodal input token vector ,$C_{hs}$ is the hidden size in the multimodal Encoder. M is then fed into the multimodal encoder and get the image feature vectors  $H \in \mathbb{R}^{C_{hs} \times N_v}$ and text feature vectors $\hat{text} \in \mathbb{R}^{C_{hs} \times N_v}$.
Finally, the $\hat{text}$ are sent to the language head for decoding, while the $H$ are sent to the image decoder for decoding (See Fig.\ref{fig:3}).

The language base model uses the pre-trained Qwen-7B model. The condition encoder implemented with a same MLP structure as the diffusion's condition encoder. By infinitely narrowing the gap between the output of the condition encoder and the output of the multimodal encoder, we can ensure the final generation quality of the decoder. Therefore, the loss function use the MSE loss(See eq.\ref{eq:6}) function to efficiently model the latent mapping between image feature $H$ and condition $C$.

\subsection{Image Decoder}
The $Img_{Dec}$ can be any model capable of performing image generation tasks. In this paper, ControlNet\cite{a31} is chosen as the $Img_{Dec}$, and its model structure can be referred to in Fig.\ref{fig:3}, Module B. In this context, the task of the image decoder is to accept a set of conditional vectors and generate images based on these vectors.

\begin{equation}
\mathcal{L} = \mathbb{E}_{\bm{m}_0, \bm{t}, \bm{c}_k, \bm{c}_\text{t}, \epsilon \sim \mathcal{N}(0, 1) }\Big[ \Vert \epsilon - \epsilon_\theta(\bm{m}_{t}, \bm{t}, \bm{c}_k, \bm{c}_\text{t})) \Vert_{2}^{2}\Big]
\label{eq:7}
\end{equation}

A segmentation mask ${m}_0  \in \mathbb{R}^{H\times W \times 3}$ as input, then pass VAE encoder and diffusion algorithms progressively introduce noise to the image, generating a latent space image ${m}_{t}  \in \mathbb{R}^{\frac{H}{8}\times \frac{W}{8} \times 4}$. Subsequently, with the keyword prompt ${c}_k  \in \mathbb{R}^{1\times 77 \times dim_p}$ and the feature vector output from the conditional model ${c}_t  \in \mathbb{R}^{\frac{H}{8}\times \frac{W}{8} \times dim_c}$ as auxiliary, a function $\epsilon_\theta$ is employed to predict noise. $dim_p$ and $dim_c$ is the the dimensions of the features.
We utilized the pre-training parameters of the SD 1.5B\cite{a30} to ensure the image decoder could be trained with relatively less data while achieving excellent image generation results. The training process can be conceptualized as the recovery of input image features from the conditioned vectors. Initially, the input image is fed into the designated conditional encoder, extracting features from the input image to generate conditional variables. Subsequently, the conditional variables are injected into the backbone network using cross-attention. The loss is calculated by comparing the noise to optimize the entire image encoder during the training process. 

\section{Experiments}
\label{sec:Exper}
\subsection{Dataset}
The training set includes 71 publicly available datasets.
Additionally, we reserved the optic fundus segmentation task to evaluate SegICL's learning capabilities for OOD tasks. 
Furthermore, for each modality, we retained some datasets to assess the performance of SegICL on OOD tasks. 
Additionally, we applied the approach described in \cite{a27} to uniformly preprocess the experimental datasets.

The evaluation perform on three optic fundus datasets, along with one CT dataset and one MRI dataset.
 \textbf{REFUGE2}\cite{a32}. 
This dataset contains 2000 color optic fundus images with annotations of glaucoma classification, optic disc/cup segmentation, as well as fovea localization. We conducted evaluations on the optic disc/cup segmentation task. \textbf{PALM}\cite{a33}. 
This dataset consists of 1200 images with labels for pathologic myopia, including manual annotations for the optic disc, fovea position. 
We chose evaluations on the optic disc segmentation task. \textbf{IDRiD}\cite{a34}. 
This dataset includes disease severity information for diabetic retinopathy and diabetic macular edema in each image. 
There are five segmentation tasks, and we chose optic disc segmentation for the evaluation experiment. \textbf{CHAOS}\cite{a35}. 
This dataset includs abdomen MRII data, which sourced from ISBI 2019 Combined Healthy Abdominal Organ Segmentation Challenge (Task 5). 
We focus on four segmentation classes:liver, left kidney, right kidney, spleen. \textbf{BTCV}\cite{a36}.
This dataset includes a total of 50 abdominal CT scans. 
We adhere to the aforementioned evaluation metrics, focusing specifically on four segmentation classes: liver, left kidney, right kidney and spleen.
\begin{figure}[tp]
  \centering
  \includegraphics[width=0.8\textwidth]{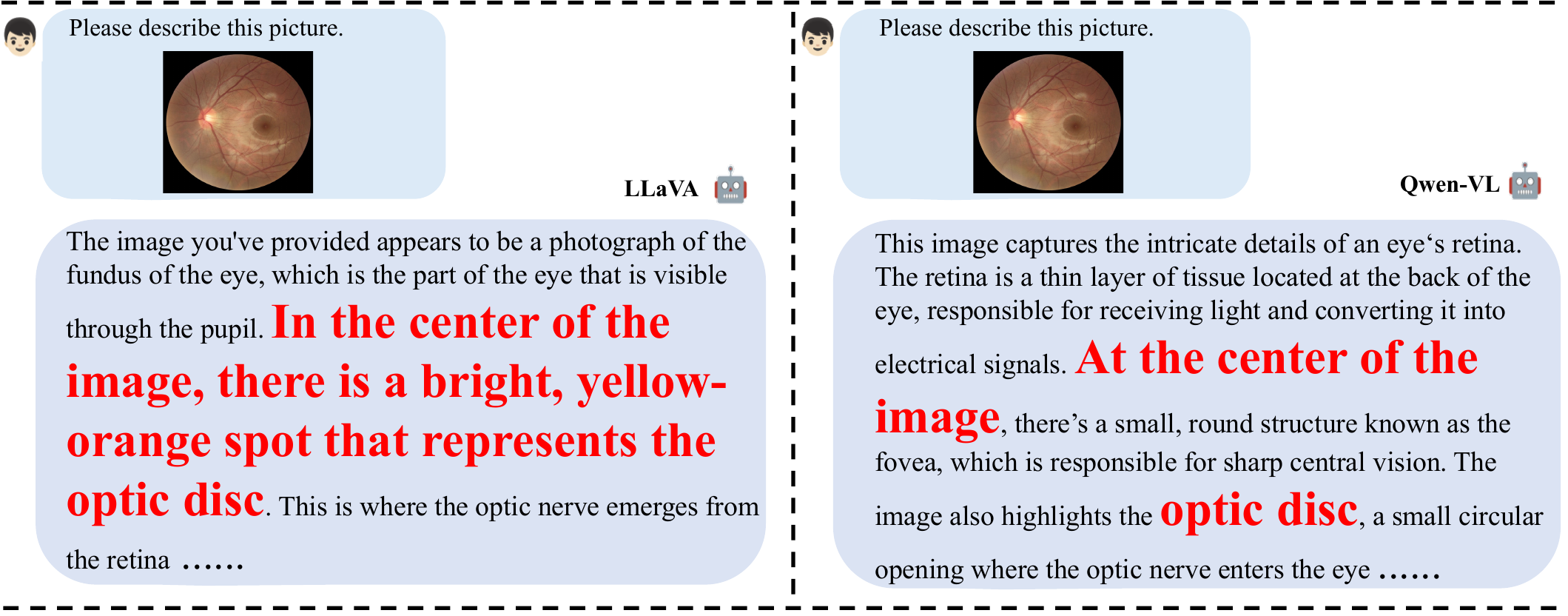}
  \caption{Diagram of the differences between different base models. Different base models possess varying degrees of prior knowledge. Thanks to pre-trained weights, the model can roughly localize the target even in a zero-shot scenario.}
  \label{fig-x2}
\end{figure}
\subsection{Implementation details}
Model training is divided into two stages (see appendix).
LoRA\cite{a44} is applied to the multimodal encoder.
The Dice\cite{a37} score is used as the evaluation metric for the model. For the calculation of few-shot performance, we randomly sample data from the training set of the dataset under evaluation as multimodal prompt. The final result is obtained by averaging the pixel-wise results of 5 independently randomly sampled predictions, followed by post-processing through binarization of the output masks.

\subsection{Performance of the OOD modality}
\begin{figure}[H]
  \centering
  \includegraphics[width=0.75\textwidth]{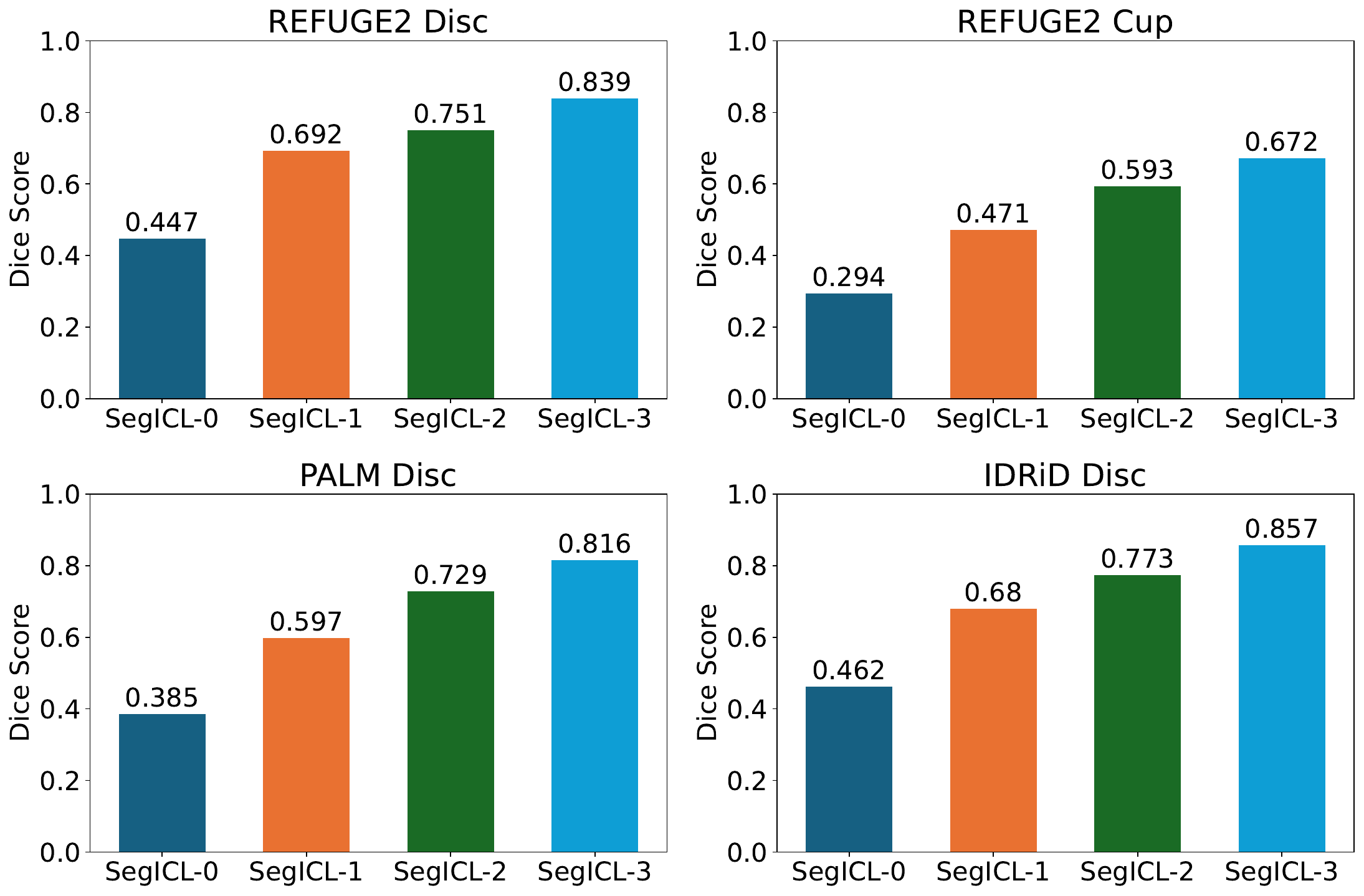}
  \caption{Performance comparison of SegICL on OOD modality. The positive correlation can be observed between the number of prompt samples (SegICL-x) and segmentation performance. Although SegICL-3 doesn't match the SOTA models, its train-free results are still adequate for assisting cold-start in semi-automatic annotation.}
  \label{fig:unseen}
\end{figure}
This section primarily analyzes modalities and their segmentation tasks beyond the distribution of the training dataset. The optic fundus image data did not utilized in the training dataset for SegICL. Pre-trained models understand optic fundus images as shown in Fig.\ref{fig-x2}, and the approximate shape and position of the optic disc can be identified. 
We evaluated the effectiveness of the proposed framework on three datasets(See Fig.\ref{fig:unseen}). On the REFUGE2 dataset, employing SegICL-0 by directly using text prompts for optic disc and cup segmentation tasks resulted in scores of 0.447 and 0.294, respectively. 
This indicates that the prior knowledge of large models can provide some help for the segmentation of the model.
\begin{figure}[H]
  \centering
  \includegraphics[width=0.85 \textwidth]{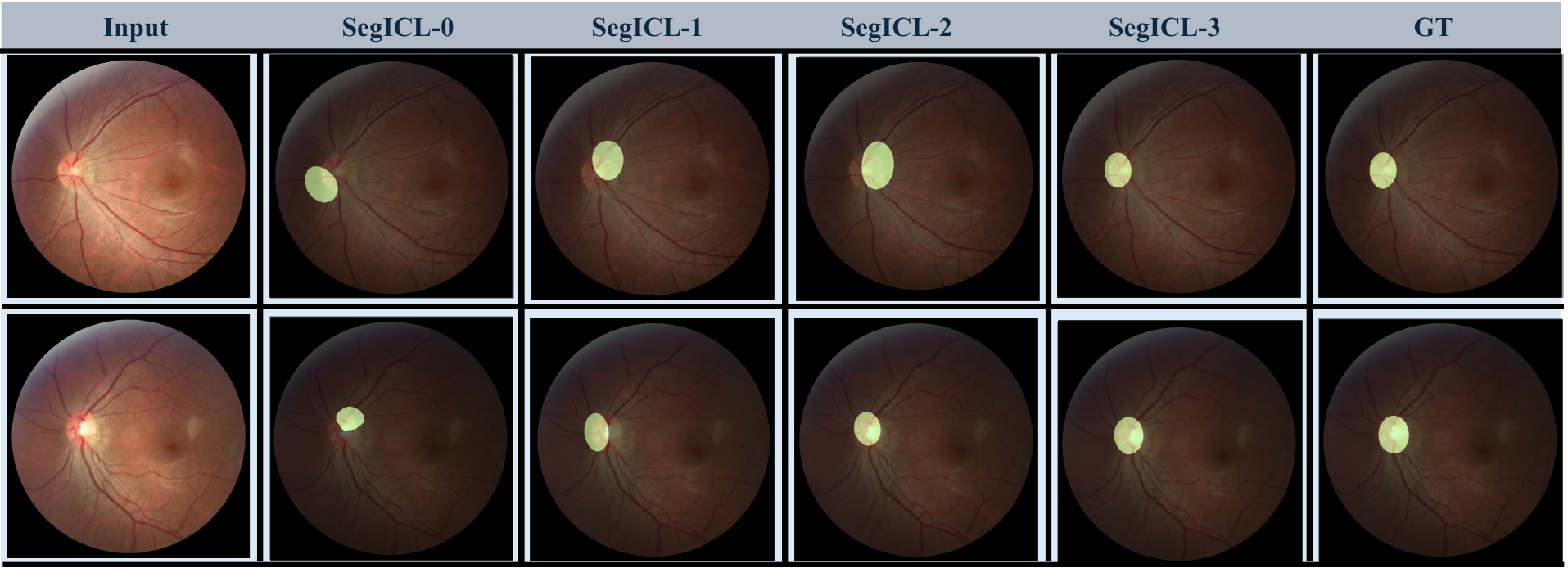}
  \caption{The qualitative analysis diagram of SegICL on optic fundus segmentation tasks. As more examples are provided, the shape and position of the masks increasingly approximate the GT.}
  \label{fig:5}
\end{figure}
For the optic disc and cup segmentation task, introducing SegICL-0 to SegICL-3 brings continuous performance growth. It can be observed that with the addition of shots, SegICL can effectively capture the feature correlations between contexts, thereby transitioning OOD tasks from being undoable to achievable albeit with suboptimal performance.
Based on the above experiments, it can be concluded that SegICL possesses powerful in-context learning capabilities. When confronted with a new task without prior training, the segmentation performance is initially low. However, through example-based teaching, the model can grasp the segmentation approach for the task (See Fig.\ref{fig:5}). With more shots provided, the model's segmentation performance on that task shows a linear growth.  

\subsection{Performance of the OOD dataset}
The generalization performance of SegICL is evaluated using an abdominal MRI image dataset that did not appear in the training dataset. 
However, despite the presence of segmentation task data similar to the training dataset in this OOD dataset, they are not sampled from the same data distribution.

\begin{table}[h]
\centering
\renewcommand\arraystretch{1.0}
\resizebox{.8\columnwidth}{!}{
\small 
\begin{tabular}{cccccccc}
\hline\hline
                Method          & Paradigm & Liver \ & Spleen \ & L.Kidney \ & R.Kidney \ & Mean    \\ 
      \hline\hline
      PANet\cite{a38}           & 1-shot   &50.40   & 40.58   &32.19      &30.99        &38.53   \\
      SE-Net\cite{a39}          & 1-shot    &29.02   & 47.30   &47.96      &45.78        &42.51   \\
      ALPNet\cite{a7}          & 1-shot   &62.35   & 61.32   &60.81      &58.83        &63.17   \\
      RPNet\cite{a18}           & 1-shot   &73.51   & 69.85   &70.00      &70.48        &79.26   \\
      GCN-DE\cite{a41}          & 1-shot    &49.47   & 60.63   &83.03      &76.07        &67.03   \\
      VQNet\cite{a17}           & 1-shot   &81.72   & \textbf{79.08}   &68.94      &60.03        &72.44   \\
      CRAPNet\cite{a51}         & 1-shot   &76.46    & 74.32    &81.95       &86.42         &79.79   \\
      \hline
      SSL-ALPNet\cite{a7}      & 1-shot   &76.10   & 72.18   &85.18      &81.92        &78.84   \\
      SSL-VQNet\cite{a17}       & 1-shot   &\textbf{79.92}   & 77.21   &91.56      &89.54        &84.56   \\
      \hline
      \textbf{}         & ICL-0    & 70.91  & 62.95   &82.20      &80.47        &73.95   \\
      \textbf{SegICL}         & ICL-1    &75.42   &71.95    &86.22      &85.90        &79.65  \\
      \textbf{}         & ICL-3    & 79.47  & 78.92   &\textbf{92.18}    &\textbf{89.95}        &\textbf{85.13}  \\
     
      \hline\hline
\end{tabular}}
\caption{ Performance comparison of SegICL and SOTA few-shot models on MRII dataset, SSL represents the utilization of self-supervised learning to enhance model performance, while SegICL-x stands for the number of samples for in-context learning.}
\label{table-MRII}
\end{table}
\begin{figure}[tb]
  \centering
  \includegraphics[width=0.68 \textwidth]{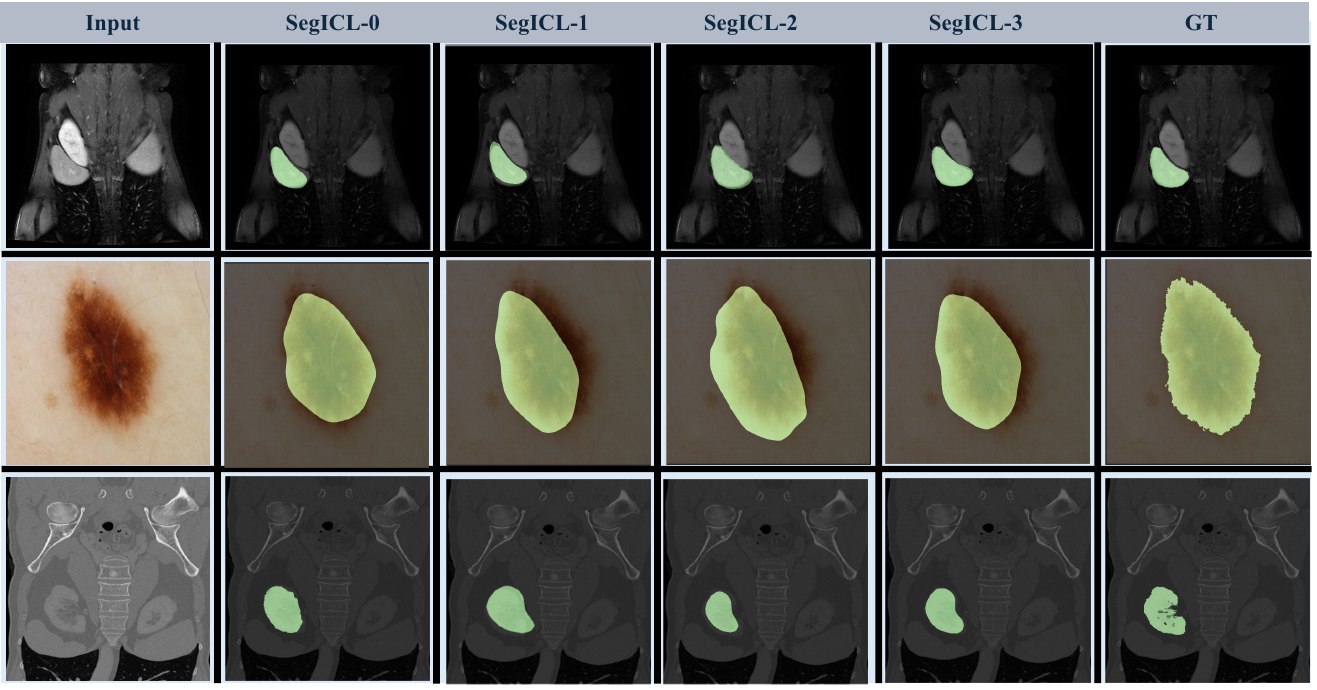}
  \caption{The qualitative analysis diagram of SegICL on OOD dataset.  Demonstrates that with an increase in provided examples, segmentation performance also improves.}
  \label{fig:6}
\end{figure}
In Tab.\ref{table-MRII}, we compared the performance of SegICL with SOTA few-shot methods. All few-shot methods are based on 1-shot settings. As shown in the table, our approach achieved an average Dice score higher than VQNet, by 1.52\% in the SegICL-0 setting. Specifically, our SegICL-0 outperformed VQNet by 13.26\% and 20.44\% in Kidney segmentation. Compared to incorporating Self-Supervised Learning (SSL), SegICL, with the introduction of In-Context Learning (ICL), demonstrated further improvement in performance. Its average Dice score surpassed the strongest few-shot method, VQNets, by 0.57\%. Moreover, it reached SOTA levels in spleen, left kidney, and right kidney sub-segmentation tasks. It is important to note that our method achieved inference in a completely train-free manner, avoiding the need to retrain the network with an SSL module.
Additionally, the table reflects the necessity of In-Context Learning for performance enhancement on OOD tasks. It is evident that, for the four sub-classification tasks, performance improved by 8.56\%, 15.97\%, 9.98\%, and 9.48\%, respectively, resulting in an average Dice improvement of 11.18\%. Thus, SegICL demonstrates robust generalization performance, achieving competitive results on OOD tasks. SegICL demonstrates a certain level of generalization ability for OOD datasets and segmentation tasks.

From Fig.\ref{fig:6}, it can be observed that the segmentation accuracy of SegICL improves with the increasing context information. The segmentation images of three different modalities are displayed, indicating that the model's contextual learning ability is effective across various modalities.
\subsection{Performance of the In-distribution datasets}
\begin{figure}[h]
  \centering
  \includegraphics[width=1\textwidth]{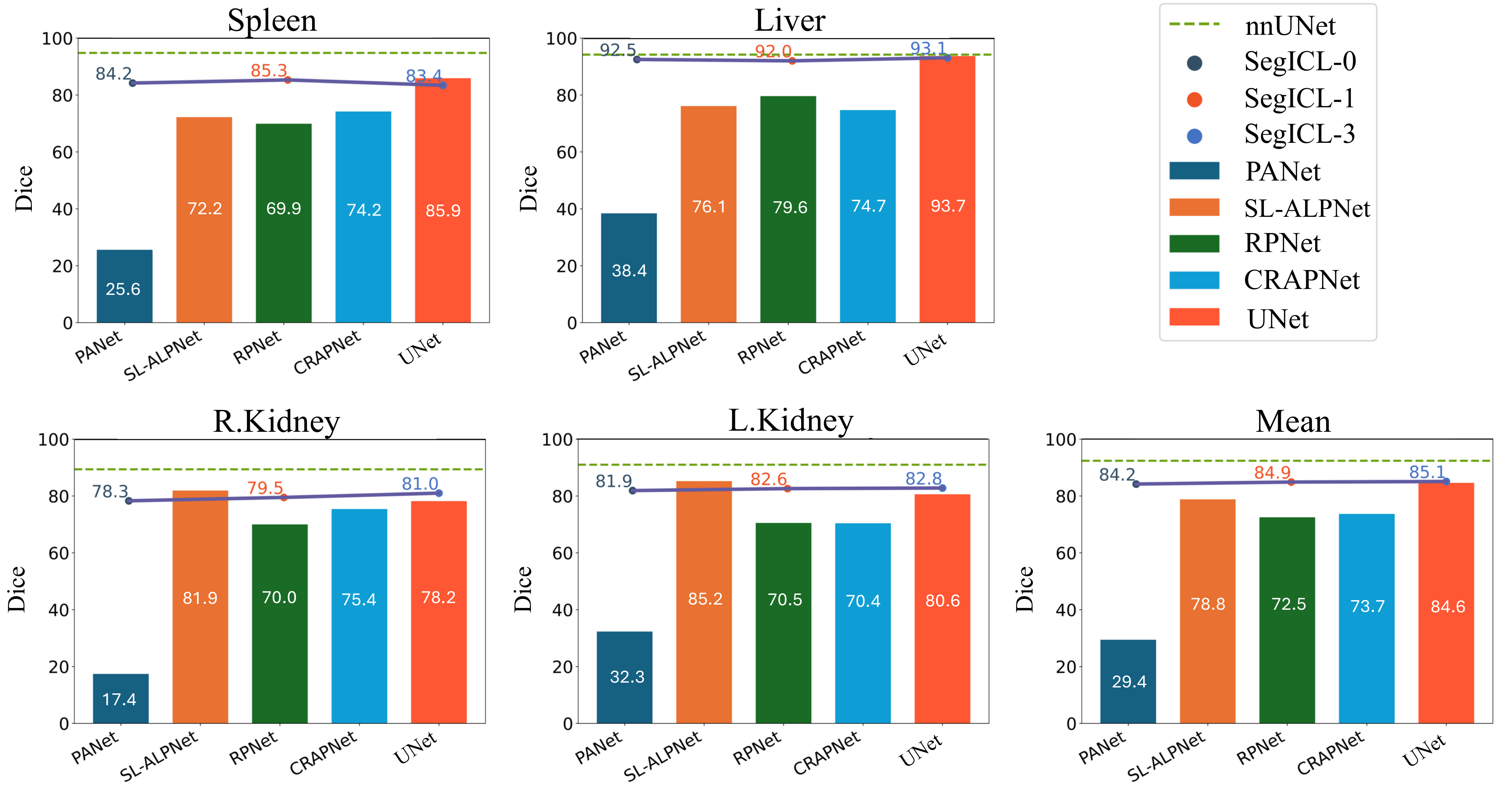}
  \caption{Performance comparison on CT dataset. The performance of SegICL is comparable to those model specifically customized for this dataset.}
  \label{fig:x}
\end{figure}
In this section, we conducted performance evaluation of SegICL using the test set from the dataset that is previously used in the training set. As shown in Fig.\ref{fig:x}, SegICL exhibits comparable performance to models trained specifically on dedicated datasets. We used nnU-Net, trained on the BTCV dataset through supervised learning, as the performance benchmark. In SegICL-0 inference, SegICL's performance differs from UNet by only 0.4\%. After employing in-context learning with example data selected from the training set, our method's average performance surpasses UNet by 0.5\%, demonstrating the learning capability of SegICL.

Simultaneously, the table indicates that in this experimental setup, the learning capacity of SegICL is noteworthy. Through in-context learning, the model is able to capture contextual features of the segmentation task, leading to improved segmentation performance.
SegICL falls short of surpassing proprietary models trained specifically for dedicated datasets due to its training dataset encompassing multiple modalities from various datasets. As a result, its performance on a single dataset can only be competitive rather than achieving SOTA status. The primary purpose behind introducing SegICL is to propose a method capable of addressing OOD tasks through ICL with a limited scale. 
\subsection{Qualitative results}
\label{sec:qr}
\begin{figure}[tb]
  \centering
  \includegraphics[width=0.75\textwidth]{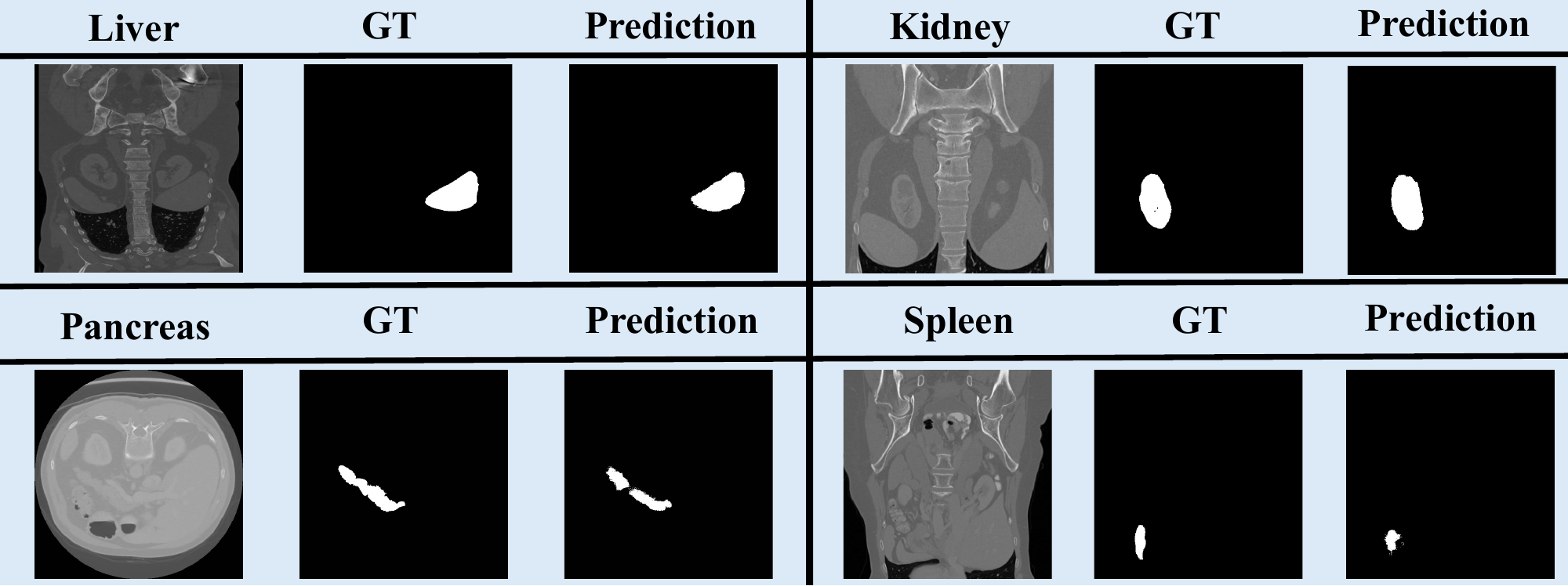}
  \caption{Qualitative analysis results display: Segmentation results for different organs in CT image. Demonstrate the strengths and limitation of SegICL.}
  \label{fig:7}
\end{figure}

For a more in-depth exploration of SegICL's segmentation performance, we visualize four typical CT images featuring the liver, spleen, kidneys, and pancreas in Fig.\ref{fig:7}. The results reveal that SegICL excels in segmenting CT slices with larger and more regular-shaped target volumes, such as the upper part displaying the liver and kidneys. However, its performance diminishes with smaller and elongated target shapes.  Qualitative analysis unveils two potential issues. First, the limited modeling capacity of the multimodal encoder for the state vector. Alternatively, increasing the number of examples can lead to better segmentation results. Secondly, the use of diffusion to generate segmentation masks inherits issues from the diffusion model, leading to a slight deficiency in generating fine-grained segmentation details. Nonetheless, existing segmentation works based on diffusion\cite{a52} have demonstrated its significant improvement in segmentation performance, maybe this issue can be solve. 
\section{Discussion and Broader Impact}
\label{sec:Discu}
\textbf{Discussion: }First, due to computing resource constraint, the token length of inputs and outputs is limited, which prevents us from exploring the performance upper limit of SegICL-x. Second, there’s still room for improvement in contour processing for the generated masks, even with post-processing, precise contour segmentation cannot be achieved. Future research can be focused on the following three aspects: Firstly, training the model using larger models and datasets, transitioning from Few-Shot to Many-Shot ICL. Secondly, improving the details of generated images is also crucial for enhancing the performance of the model. Thirdly, accelerating the segmentation speed of the model. Due to the introduction of large language models and diffusion models, the segmentation speed is relatively slow, so it is necessary to study how to improve the speed of the model.

\textbf{Broader Impact: }In this paper, we introduce SegICL, a pioneering approach to image segmentation leveraging text-guided segmentation and ICL that effectively addresses the limitations of universal segmentation models on OOD task with a training-free manner. Our work redefine the paradigms for  segmentation tasks in the era of larger models and paves the way for more intuitive, efficient, and effective segmentation models that can adapt to the rapidly evolving landscape of medical image.

\bibliography{main}
\bibliographystyle{plain}
\newpage
\end{document}